\documentclass[10pt,twocolumn,letterpaper]{article}

\usepackage[pagenumbers]{iccv} 
\usepackage{booktabs}       
\usepackage{amsfonts}       
\usepackage{nicefrac}       
\usepackage{microtype}      
\usepackage{graphicx}
\usepackage{array}
\usepackage{wrapfig}
\usepackage{multirow}
\usepackage{colortbl}
\usepackage{makecell} 
\usepackage{amsmath}
\usepackage[ruled,vlined]{algorithm2e}
\usepackage{cancel}
\definecolor{iccvblue}{rgb}{0.21,0.49,0.74}
\usepackage[pagebackref,breaklinks,colorlinks,allcolors=iccvblue]{hyperref}

\title{PathVQ: Reforming Computational Pathology Foundation Model for\\ Whole Slide Image Analysis via Vector Quantization}

\author{%
  Honglin Li$^{1,3}$ \space Zhongyi Shui$^{1,3}$  \space Yunlong Zhang$^{1,3}$ 
  Chenglu Zhu$^{2,3}$\thanks{Corresponding Author} \space \space Lin Yang$^{2,3*}$\\
  $^{1}$ Zhejiang University\\
  $^{2}$ Research Center for Industries of the Future and $^{3}$ School of Engineering, Westlake University\\
  \texttt{\{lihonglin,zhuchenglu,yanglin\}@westlake.edu.cn} \\
}
\begin{document}
\maketitle
\begin{abstract}
Computational pathology and whole-slide image (WSI) analysis are pivotal in cancer diagnosis and prognosis. However, the ultra-high resolution of WSIs presents significant modeling challenges. Recent advancements in pathology foundation models have improved performance, yet most approaches rely on [CLS] token representation of tile ViT as slide-level inputs (16×16 pixels is refereed as patch and 224×224 pixels as tile). This discards critical spatial details from patch tokens, limiting downstream WSI analysis tasks.
We find that leveraging all spatial patch tokens benefits WSI analysis but incurs nearly 200× higher storage and training costs (e.g., 196 tokens in ViT$_{224}$). To address this, we introduce vector quantized (VQ) distillation on patch feature, which efficiently compresses spatial patch tokens using discrete indices and a decoder. Our method reduces token dimensionality from 1024 to 16, achieving a 64× compression rate while preserving reconstruction fidelity.
Furthermore, we employ a multi-scale VQ (MSVQ) strategy, which not only enhances VQ reconstruction performance but also serves as a Self-supervised Learning (SSL) supervision for a seamless slide-level pretraining objective. 
Built upon the quantized patch features and supervision targets of tile via MSVQ, we develop a progressive convolutional module and slide-level SSL to extract representations with rich spatial-information for downstream WSI tasks.
Extensive evaluations on multiple datasets demonstrate the effectiveness of our approach, achieving state-of-the-art performance in WSI analysis. Code will be available soon.
\end{abstract}    
\section{Introduction}
\label{sec:intro}

\begin{figure}[htbp]
\centering
\includegraphics[width=1.\linewidth]{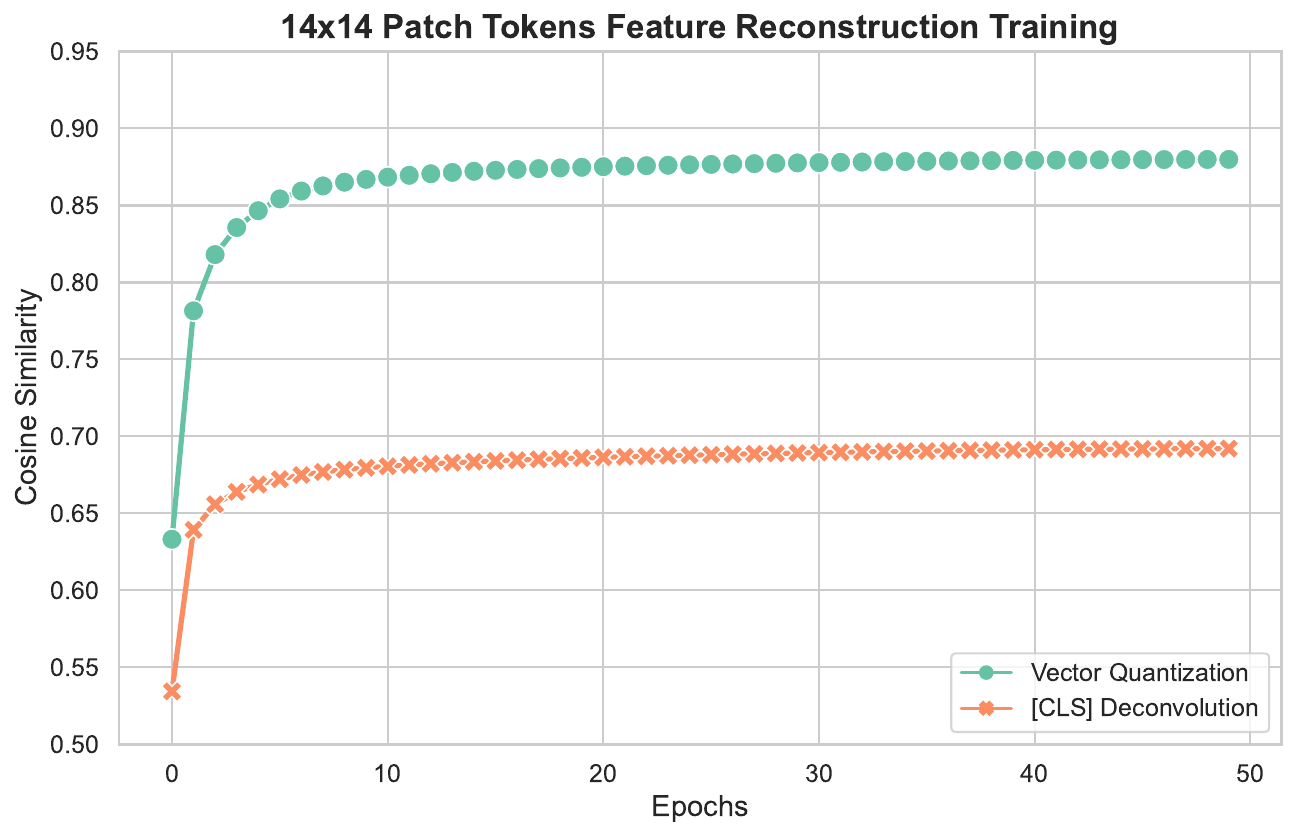} \\
\caption{Evaluation on information loss via reconstruction training. Directly using the [CLS] token results in significant information loss, making it difficult to reconstruct all patch tokens, potentially discarding critical details for downstream tasks. In contrast, vector quantization retains more original information.}
\label{fig1}
\vspace{-4mm}
\end{figure}

Cancer remains one of the most challenging diseases to diagnose and prognosticate, with pathology playing a pivotal role in understanding its complexities~\cite{huang2020artificial}. Traditional histopathological analysis relies heavily on manual examination of tissue samples by pathologists, a process that is not only time-consuming but also prone to inter-observer variability~\cite{gurcan2009histopathological}. In recent years, computational pathology (CPath) has emerged as a transformative field, leveraging whole-slide images (WSIs) to enable automated and quantitative analysis of tissue samples~\cite{lu2021data,xu2024whole,lu2024visual}. WSIs, which are high-resolution digital scans of entire tissue slides, provide a wealth of information that can be harnessed for cancer diagnosis, prognosis, and treatment planning. However, the ultra-high resolution of WSIs, often exceeding billions of pixels, presents significant challenges for effective computational modeling~\cite{transmil,li2024rethinking}.

Recent advancements in AI~\cite{openai2023gpt,touvron2023llama} and foundation models~\cite{bommasani2021opportunities,dino} have shown remarkable promise in addressing these challenges. Foundation models, pre-trained on large-scale datasets, have demonstrated exceptional performance in various computer vision tasks~\cite{kirillov2023segment,oquab2023dinov2}, including WSI analysis~\cite{xu2024whole,lu2024visual,chen2024towards}. These computational pathology foundation models (CPathFMs) typically process WSIs by dividing them into smaller tiles (e.g., 224$\times$224 pixels as a tile), extracting features from each tile, and aggregating these features to make slide-level predictions. However, most existing approaches rely on [CLS] token representations of tile-level ViT~\cite{ViT} as slide-level inputs, where 16$\times$16 pixels are referred to as a patch. This approach discards critical spatial details from patch tokens, limiting the performance of downstream WSI analysis tasks.
Notably, some studies have attempted to address this issue by scaling up to larger models (UNI-2~\cite{chen2024towards}) or combining [average pooling] features with the [CLS] token (Virchow-2~\cite{zimmermann2024virchow2}). However, these approaches yield only marginal improvements. Consequently, we argue that the performance of CPathFMs is fundamentally constrained by information loss associated with the [CLS] token.

Leveraging all spatial patch tokens may benefit WSI analysis but incurs nearly 200$\times$ higher storage and training costs (e.g., 196 tokens in ViT$_{224}$). To address this, we introduce {vector quantization~\cite{van2017vqvae,peng2022beit} (VQ) distillation} on patch features, which efficiently compresses spatial patch tokens using discrete indices and a decoder. Our method reduces token dimensionality from 1024 to 16, achieving a 64$\times$ compression rate while preserving reconstruction fidelity. This compression process retains most spatial and contextual information, ensuring that critical features are preserved for downstream tasks.

Furthermore, we employ a multi-scale VQ (MSVQ) strategy, which unifies patch-level feature VQ and tile-level feature VQ. Intuitively, tile-level feature like [CLS] token can be seen adaptive combination all patch features, thus they share the same feature space and can be learned into a single VQ model. This not only enhances VQ reconstruction performance but also serves as a Self-supervised Learning (SSL) supervision target for a seamless slide-level pretraining objective (working as a tokenizer thus can be pretrained like BERT~\cite{BERT,MAE,peng2022beit}). 
By integrating SSL into our framework, we enable the model to learn rich, discriminative representations from unlabeled WSIs, addressing the challenge of limited WSI samples in computational pathology downstream tasks. 
Built upon the quantized features of patches and supervision targets of tiles via MSVQ, we develop a progressive convolutional module and slide-level SSL to extract representations with rich spatial information for downstream WSI tasks, leading to more accurate and interpretable predictions for cancer diagnosis and prognosis.

The contributions of our work can be summarized as follows:
\begin{itemize}
    \item Efficient Token Compression with VQ Distillation: We propose a novel VQ-based framework that compresses patch-level spatial tokens by 64$\times$ while retaining critical spatial and contextual information, enabling scalable and efficient WSI analysis.
    \item SSL Supervision via offline tokenizer: Our improved MSVQ strategy not only enhances feature reconstruction but also serves as an SSL supervision target for slide-level mask prediction, providing a new direction for pretraining WSI models.
    \item Rigorous Validation: Extensive evaluations on multiple datasets demonstrate the effectiveness of our approach, achieving state-of-the-art performance in WSI analysis tasks, with practical implications for clinical applications.
\end{itemize}

By addressing the computational challenges of WSI analysis while preserving critical spatial information, our framework offers a new perspective on the development of computational pathology foundation models, paving the way for more accurate and scalable cancer diagnostics.

\section{Related Work}
\label{sec:rel}

\subsection{Pathological Whole Slide Image Analysis}
Whole Slide Images (WSIs) contain a wealth of visual information that plays a crucial role in pathological analysis~\cite{campanella2019clinical,lu2021data}. However, obtaining detailed cell-level annotations is both labor-intensive and time-consuming~\cite{campanella2019clinical,lu2021data,chen2022scaling}, posing a significant challenge for large-scale WSI analysis. To address this, weakly-supervised learning has emerged as a promising direction in computational pathology.

\noindent \textbf{Computational Pathology Foundation Models}
The performance of early WSI MIL approaches~\cite{abmil,chen2021slide,li2021dual,lu2021data,Zhang2022DTFDMILDF,javed2022additive,qu2022bidirectional,cui2023bayesmil,bergner2023iterative,chen2022scaling,transmil,li2024rethinking,chen2021slide,li2018graph,guan2022node,chan2023histopathology,fourkioti2024camil} relied heavily on tile-level features extracted from pre-trained models~\cite{Li_2023_CVPR,Filiot2023phikon}. To address this, CPathFMs have been developed and shown significant advancements in both tile-level~\cite{wang2021transpath,lu2024visual,chen2024towards,Filiot2023phikon,vorontsov2023virchow,hoptimus0} and WSI-level analysis~\cite{xu2024whole,ding2024titan}. These CPathFMs leverage visual Self-supervised Learning (SSL) techniques~\cite{dino,oquab2023dinov2,MOCOV3,chen2020simple} on large-scale unlabeled datasets~\cite{wang2021transpath,chen2024towards,Filiot2023phikon} or organize pathology image-text pairs to learn multimodal representation~\cite{CLIP,ding2024titan,lu2024visual}. CPathFMs have demonstrated superior performance in downstream tasks such as cancer subtyping, survival prediction, and biomarker identification.

Recently, authors in Hest-1k~\cite{jaume2025hest} observe that tile encoders like Conch~\cite{lu2024visual} can be further fine-tuned to
obtain better downstream tile task result.
However, the key challenge that the computational cost high-resolution WSIs makes fine-tuning ~\cite{zhang2023promptmil,Li_2023_CVPR} CPathFMs with overwhelm parameters difficult. Most approaches~\cite{lu2024visual,chen2024towards,Filiot2023phikon} resort to using pretrained [CLS] token representation of tile-level CPathFM as slide-level inputs, which may lead to the loss of critical spatial information. Some models, such as UNI-2~\cite{chen2024towards}, attempt to scale up ViTs into larger-size as tile encoders to extract better feature representations, but only achieve marginal improvements from ViT-Large to ViT-Giant. We argue that this performance bottleneck stems from the spatial information loss inherent in [CLS] token representations. Other efforts, such as Virchow-2~\cite{zimmermann2024virchow2}, find that combining [CLS] tokens with [AVG] (average pooling of all spatial tokens) can yield some improvement (less than 1 point).
Motivated by these findings, we propose to keep but compress all spatial patch tokens and further extract useful information for downstream WSI tasks analysis.

\noindent \textbf{Slide-level CPathFMs / SSL pretraining:}
Some recent slide-level CPathFMs, e.g. GigaPath ~\cite{xu2024whole} and TITAN ~\cite{ding2024titan} are modeled via Transformer with 6 to 12 layers, then pretrained via MAE~\cite{MAE} and iBOT~\cite{zhou2021ibot} respectively. Some other slide SSL models~\cite{hou2024self,lazard2023gigassl,chen2022scaling} also propose to employ slide-level augmentation with contrastive learning (CL) ~\cite{chen2020simple,oquab2023dinov2} for pretraining. 
But there are some training problems of these works:
1) The main augmentations in slide-level to generate different views for CL is limited, like crop or random-drop, since the tile feature are pre-processed and stored. This hinders the performance of CL pretraining.
2) The self-supervised target. The iBOT, ~\cite{zhou2021ibot} used in TITAN~\cite{ding2024titan} predicting masked token to match online-tokenizer, is not so stable during training. The MAE in GigaPath~\cite{xu2024whole} need to regress the feature of masked tiles, which may be too difficult to fit and hinder downstream tasks. The CHIEF~\cite{wang2024pathology}, on the other way, pretrain ABMIL by constrastively predicting tumor organ source (extra information).

Different to these work, in this paper we train a offline tile tokenizer via vector quantization which can offer self-supervision for both WSI-Transformer mask modeling and ABMIL. This is more stable for pretraining and need no further information.

\subsection{Vector Quantization}
Vector quantization (VQ)~\cite{gray1984vector} is a fundamental technique in signal processing and machine learning, widely used for data compression, clustering, and generative modeling~\cite{equitz1989new,van2017vqvae,razavi2019vq_vae2}. Recent developments in deep learning have led to neural vector quantization methods, such as Vector Quantized Variational Autoencoders (VQ-VAEs)~\cite{van2017vqvae,razavi2019vq_vae2,chang2022maskgit} and quantized transformers~\cite{lingle2023transformer}, which integrate VQ into end-to-end learning pipelines to enhance expressiveness and efficiency. To further improve VQ, techniques such as residual quantization~\cite{lee2022autoregressive} and rotation tricks~\cite{fifty2025restructuring} have been proposed. Recent studies~\cite{yu2021vector,yu2023language} reveal that lower-dimensional quantized vectors (dimension size ranging from 8 to 32) can improve codebook usage and reconstruction performance, providing strong compression capabilities that benefit this study. Unlike recent VQ methods that focus primarily on visual generation~\cite{van2017vqvae,tian2025var,razavi2019vq_vae2}, our work focus on feature compression and distillation of pretrained pathology tile encoder features via a VQ quantizer.
\section{Method}
\label{sec:method}

\begin{figure*}[htbp]
\centering
\includegraphics[width=1.\linewidth]{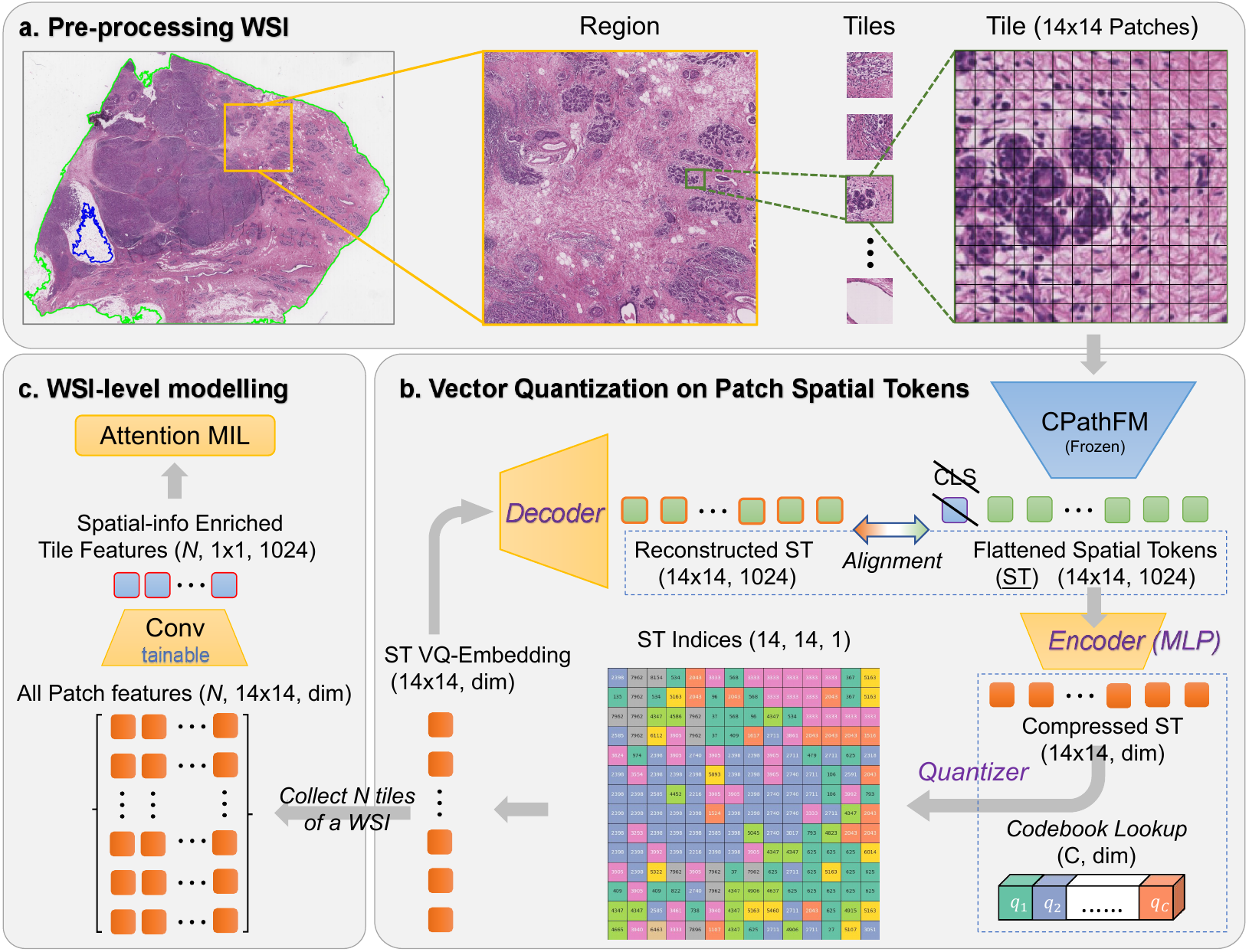} \\
\caption{Overview of the proposed framework. (a) The pipeline for compressing spatial patch tokens using vector quantization (VQ) and multi-scale VQ (MSVQ). (b) Slide-level self-supervised learning (SSL) using MSVQ-generated tokenizers. (c) Downstream WSI task fine-tuning with compressed patch features.}
\label{fig_framework}
\end{figure*}

\subsection{Preliminary}
\label{sec:preliminary}

Given a pathology tile image $x$, the pretrained CPathFM ViT first converts it into $n$ patches: $x_{tile} = [p_1, p_2, \dots , p_n]$, where the most commonly used patch size is $16 \times 16$. The ViT outputs a representation: $x_o = [s; h_1, h_2, \dots, h_n]$, where $s$ serves as a summary [CLS] of the spatial tokens of all patches ($\textbf{ST} = [h_1, h_2, \dots, h_n]$). 

For WSI modeling, a WSI $X$ is first divided into $N$ tiles: $X=[x_1, x_2,\dots, x_N]$, which are then processed by the CPathFM ViT as above. Most existing approaches rely on the [CLS] token from each tile as the slide-level representation, forming tile embeddings $S=[s_1, s_2, \dots, s_N] \in \mathbb{R}^{N\times D}$. These embeddings are subsequently aggregated for slide-level prediction: $\hat{Y} = g(S; \theta)$, where $g(\theta)$ can be an attention mechanism or a Transformer. 

In contrast, this paper explores using all spatial tokens $H=[\textbf{ST}_1,\textbf{ST}_2,\dots,\textbf{ST}_N] \in \mathbb{R}^{N\times n \times D}$ for slide-level prediction. Here, $N$ (the number of instances) can exceed 10k, $n=196$ (the number of patches per tile), and the representation dimension $D=1024$ (for UNI~\cite{chen2024towards}). So, directly leveraging these high-dimensional features is computationally prohibitive for WSI training.

\subsection{Vector Quantization Learning}
To mitigate the computational burden while incorporating all patches' ST representations, we introduce vector-quantization (VQ) learning on the pretrained CPathFM's patch ST, as illustrated in Figure~\ref{fig_framework}b. This framework consists of an encoder, a quantizer, and a decoder. Additionally, we extend VQ to support both patch and tile representations via a multi-scale VQ strategy.

\subsubsection{VQ for Patches}
\label{sec:vq_patch}
The spatial tokens (ST) are mapped into discrete codes through VQ. Specifically, the tile-level ST representation $\textbf{ST} = [h_1, h_2, \dots, h_n ]$ is first processed by an MLP encoder to reduce its dimensionality from $D$ to $d$:
\begin{equation}
    [e_1, e_2, \dots, e_n] = \text{Enc} ([h_1, h_2, \dots, h_n ]),
\end{equation}
where the reduced representations $[e_1, e_2, \dots, e_n]$ are then tokenized into discrete indices $\textbf{ST}_{tok} = [z_1, z_2, \dots, z_n]$. The employed codebook $V=[v_1, v_2, \cdots, v_{C}]\in \mathbb{R}^{C\times d}$ contains $C$ learnable codebook embeddings. The vector quantizer assigns each patch representation $e_i$ to its closest entry in the codebook:

\begin{equation}
    z_i = \arg \min_{j} || \ell_2( e_i ) - \ell_2( v_j )||_2, 
\label{eq:vq_distance}
\end{equation}
where $j \in \{1, 2, \dots, C\}$, and $\ell_2$ normalization is applied for codebook lookup, ensuring that each token is matched to the most similar vector in the codebook.

After quantization, the embeddings $E_{z_0}, E_{z_1},\dots,E_{z_n}$ are passed to a multi-layer Transformer decoder, which reconstructs the original ST representation. During training, the decoder output $o_i$ is aligned with the original ST target $h_i$ by maximizing their cosine similarity.

Since the quantization process in Equation~\ref{eq:vq_distance} is non-differentiable, the straight-through gradient estimator~\citep{van2017vqvae} is adopted, where gradients are copied from the decoder input to the MLP encoder output for backpropagation.

The training objective of VQ is formulated as:
\begin{equation}
\mathrm{max} \sum_{x \in D} \sum_{i=1}^N
\cos{( o_i , h_i )}
- ||\ell_2( e_i ) - \ell_2( v_i )||_2^2,
\label{eq:vq_objective}
\end{equation}
where $D$ denotes the dataset of image tiles. For clarity, we omit the straight-through gradients and stop-gradient operation~\citep{van2017vqvae}. During training, the MLP encoder, codebook embeddings, and decoder are updated to optimize reconstruction.

\subsubsection{Multi-Scale Vector Quantization}
\label{sec:msvq}
To simultaneously compress patch ST representations and generate an offline tokenizer for WSI self-supervised learning (SSL), we introduce a Multi-Scale VQ (MSVQ) model. MSVQ encodes CPathFM tile ST representations into $K$ multi-scale discrete token maps $R=(r_1, r_2, \dots, r_K)$.

MSVQ follows the same architecture as Section \ref{sec:vq_patch} but incorporates a multi-scale quantization module. The encoding process is designed with residual connections~\cite{tian2025var,lee2022autoregressive}, as outlined in Algorithm~\ref{alg:msvq}. 

\begin{algorithm}[h]
    \caption{\small{Multi-Scale VQ Encoding}} \label{alg:msvq}
    \small{
    \textbf{Inputs: } CPathFM's spatial token feature $ST = [h_1, h_2, \dots, h_n ]$\;
    \textbf{Hyperparameters: } steps $K$, resolutions $(H_k,W_k)_{k=1}^{K}$\;
    $f = \text{Enc}(ST)$, $R=[]$\;
    \For {$k=1,\cdots,K$}
    {
    $r_k = \mathcal{Q}(\text{interpolate}(f, H_k, W_k))$\;
    $R = \text{queue\_push}(R, r_k)$\;
    $z_k = \text{lookup}(Z, r_k)$\;
    $z_k = \text{interpolate}(z_k, H_k, W_k)$\;
    $f = f - \phi_k(z_k)$\;
    }
    \textbf{Return: } multi-scale tokens $R$ and indices $Z=[z_0,z_1,\dots,z_K]$ \;
    }
\end{algorithm}

\begin{figure}[htbp]
\centering
\includegraphics[width=1.0\linewidth]{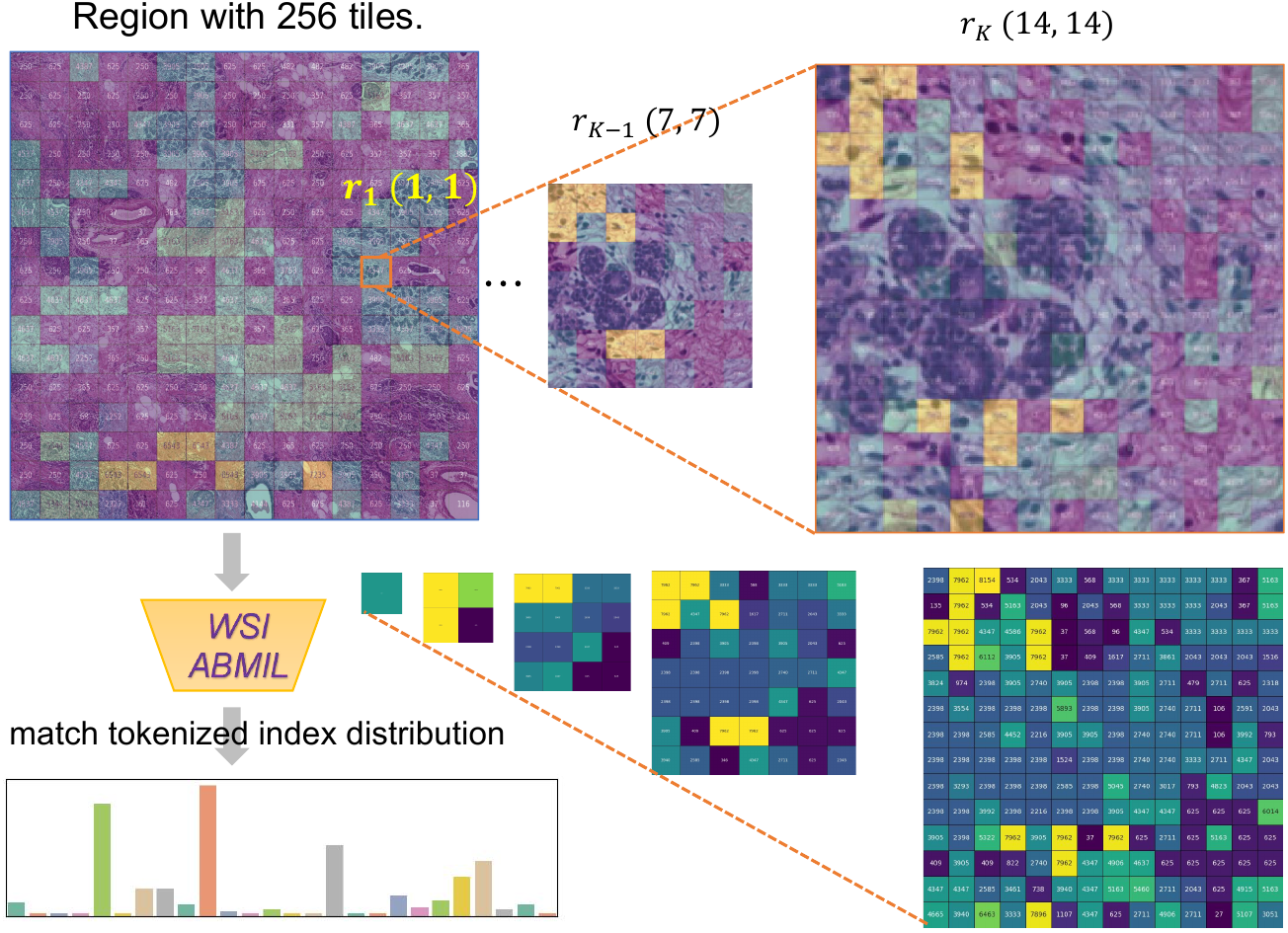} \\
\caption{Multi-scale Vector Quantization (MSVQ) visualization. Based on MSVQ, the tile- and patch-level can be quantified simultaneously for slide-level pretraining and feature compression, respectively. The region data can be used to pretrain ABMIL via token index frequency matching.}
\label{fig:msvq}
\end{figure}

Intuitively, if $R=(r_1)$, MSVQ reduces to a simple VQ applied to the average-pooled representation of all patch tokens, analogous to the tile-level [CLS] representation. Conversely, if $R=(r_K)$, MSVQ operates identically to the patch-level VQ in Section~\ref{sec:vq_patch}. The generalized form $R=(r_1, \dots, r_K)$ enables VQ at both the tile and patch levels, as well as intermediate scales. Please check Figure \ref{fig:msvq} for more visualization.

A shared codebook $Z$ is utilized across all scales, ensuring that each $r_k$'s tokens are drawn from the same vocabulary $[V_C]$. The decoding process follows the same principles but in reverse.

\subsection{Slide-Level Self-Supervised Learning}
\label{sec:ssl}

Leveraging the offline tokenizer provided by MSVQ for all WSI tiles, we design a self-supervised learning (SSL) pretraining framework for WSI-MIL analysis. This is applicable to both mainstream architectures: ABMIL and Transformer-based models.

\subsubsection{ABMIL-Based SSL}
In supervised ABMIL learning (e.g., WSI classification), adaptive pooling (or max-pooling) is commonly applied to aggregate tile instances to match the WSI-level label. Inspired by this, we propose a simple SSL objective for ABMIL based on MSVQ’s level-1 quantized indices.

Given a large-region crop from WSIs (e.g., region resolution $3584 \times 3584$, which corresponds to $[16 \times 224] \times [16 \times 224]$ and contains 256 tiles), we define the SSL training objective for each region $x$ as:
\begin{equation}
    \mathcal{L}(\theta) = -\sum_{c=1}^{C} q_c(x) \log p_\theta(x)_c,
    \label{ssl_loss:abmil}
\end{equation}
where the soft target \( q_c(x) \) is derived from the token frequency distribution over the 256 tiles, computed using the MSVQ tokenized index $z$ with $C$ categories. Specifically, \( q_c(x) \) represents the normalized frequency of tokens belonging to class \( c \) within the region. 

The predicted probability \( p_\theta(x)_c \) for each class \( c \) is obtained by processing the region through an ABMIL model, followed by a classifier layer with softmax activation:
\begin{equation}
p_\theta(x)_c = \text{softmax}\{\text{classifier}[\text{AttnPool}(x)]\}_c.
\end{equation}

\subsubsection{WSI-Transformer-Based SSL}
We adopt a masked image modeling (MIM) strategy similar to MAE~\cite{MAE} and BEIT~\cite{peng2022beit}. However, unlike standard MIM, where raw image patches serve as input, our pretraining input consists of pre-extracted feature representations.

Given an input region consisting of $k$ tiles, represented as $x = \{ t_1, t_2, ..., t_k \}$, we randomly mask a portion of tiles. The masked tile positions are denoted as $M$. A shared learnable embedding $e_{[\text{M}]}$ replaces the masked tile embeddings, and Rotary Positional Embedding (RoPE)~\cite{su2021roformer} is applied to retain spatial coherence. The corrupted region representation is:
\begin{equation}
x_{\text{corrupt}} = \{ t_1, \cancel{t_2}, ..., \cancel{t_i}, t_{i+1},..., t_k\}.
\end{equation}

For each masked tile, a softmax classifier predicts the corresponding tokenized index, which is obtained from the level-0 quantized index of the MSVQ tokenizer (see Section~\ref{sec:msvq}). This provides stable supervision signals for the masked modeling SSL procedure.

The training objective is defined as:
\begin{equation}
\mathcal{L}_{\rm{mask-modeling}}(\theta) = - \sum_{x \in D} \sum_{i \in M} \log p( z_i | x^{M}_i ),
\label{eq:mim_objective}
\end{equation}
where $z_i$ represents the tokenized index of each tile, and $D$ is the set of all pretraining region data.
Compared to iBOT’s online tokenizer, our MSVQ-based offline tokenizer provides a more stable supervision signal for SSL pretraining.

\subsection{WSI Downstream-Task Fine-Tuning}

As illustrated in Figure~\ref{fig_framework}c, the refined WSI input consists of patch feature embeddings with a compressed shape of $(N,14,14,\text{dim})$, where $N$ represents the total number of tiles in a WSI, and $(14,14)$ corresponds to the standard $2$D patch arrangement in ViT for each tile.
To enhance the feature representation for downstream tasks, we first apply convolutional layers (Convs) with a stride greater than $1$, increasing the output channel size to better capture task-relevant information. The extracted features are then reshaped to match the original CLS token representation of tile-based ViT. 

It is notable that the encoding direction of ST is not so controllable since the the encoded embeddings are in the middle layers (after encoding, before decoding). To keep its feature space, we align the output of Convs to original level-0 tile feature during VQ pre-training. This also alleviate the random initialization problem of Convs. This module will get further fine-tuned during slide-level task. We show some ablations on this issue in experiment Section \ref{sec:ablation}.
Finally, the processed features are fed into downstream Attention-based MIL models, including both ABMIL and WSI-Transformer (see Section~\ref{sec:preliminary}).


\subsection{Overall Framework and Implementation}

We summarize the overall framework of our method below. The WSI pre-processing follows the approach used in previous work~\cite{lu2021data}:

\begin{itemize}
    \item \textbf{Patch-Level VQ Learning (Figure~\ref{fig_framework}b)}: This module aims to compress all patch token features from CPathFMs, making them trainable for downstream tasks. Multi-scale VQ learning (Figure~\ref{fig:msvq}) further enables slide-level SSL supervision and subsumes patch-level VQ learning.
    
    \item \textbf{Slide-Level SSL (Figure~\ref{fig:msvq})}: Leveraging the tile-level tokenizer learned via MSVQ, SSL can be effectively applied to both ABMIL and WSI-Transformer models.
    
    \item \textbf{WSI Downstream-Task Fine-Tuning (Figure~\ref{fig_framework}c)}: Fine-tuning serves two purposes:  
    (a) transforming patch features into a more suitable representation for downstream tasks, and  
    (b) fine-tuning the pretrained slide-level SSL model for improved performance.
\end{itemize}

\section{Experiments}

In this section, we evaluate the performance of the proposed method and compare it with various baselines. Additionally, we conduct ablation studies to further analyze its effectiveness.

\subsection{Pretraining Implementation Details}

\noindent \textbf{VQ Pretraining:}  
We conduct VQ pretraining on 1M randomly cropped $224 \times 224$ tiles extracted from all TCGA~\cite{tcga} diagnostic pathology WSIs.  

During training, the CPathFM backbone (e.g., UNI with ViT-Large) remains frozen. 
The codebook has a size of $C=8192$ with an embedding dimension of $16$.

For MSVQ, we employ a multi-scale resolution list: $\{1\times1, 2\times2, 4\times4, 7\times7, 14\times14\}$.  

The model is trained on 4 RTX-3090 GPUs for 50 epochs using a batch size of 128 tile images. The total training time is approximately 22 hours.

\noindent \textbf{WSI-SSL Pretraining:}  
We crop all TCGA diagnostic WSIs into regions of resolution $3584 \times 3584$, yielding a dataset of approximately 250k regions. To facilitate SSL, a pretrained MSVQ model is used to extract the quantized indices of each tile within a region, requiring only about 65MB for storage.

During pretraining, the indices of each region are first re-embedded via a frozen VQ module, resulting in a feature representation of shape $(256, 14\times14, 16)$. The convolutional module consists of four conv layers with a stride of 2, progressively increasing the output channels from $128 \rightarrow 256 \rightarrow 512 \rightarrow 1024$. This process transforms the features into spatially enriched embeddings of shape $(256, 1\times1, 1024)$. These embeddings are subsequently fed into either an ABMIL model or a 6-layer WSI-Transformer for pretraining.



\subsection{Downstream Tasks}

We primarily focus on WSI classification and survival prediction. For dataset details, please refer to Appendix. For illustration purpose, we also run two experiments on ROI classification to clarify [CLS] token is not all we need.

The data processing and embedding procedure are identical to the region-based approach but are applied at the WSI level. The $(x, y)$ coordinates of tiles are also stored to facilitate positional encoding in the WSI-Transformer.  

During fine-tuning, both the convolutional module and the WSI model are trained with a batch size of 1 for 20 epochs. The learning rate is fixed at $1\times10^{-4}$, with a weight decay of $1\times10^{-4}$, using the AdamW optimizer with default settings.  

For ABMIL, both randomly initialized and pretrained models are fine-tuned using the same hyper-parameters and training protocol.  
For WSI-Transformer, LoRA~\cite{hu2021lora} adaptation (applied to all \texttt{nn.Linear} layers) is used with $rank=16$ during fine-tuning of the pretrained model to mitigate overfitting. For Transformer initialized from scratch, full fine-tuning is employed.  

\begin{figure}[htbp]
\centering
\includegraphics[width=1.\linewidth]{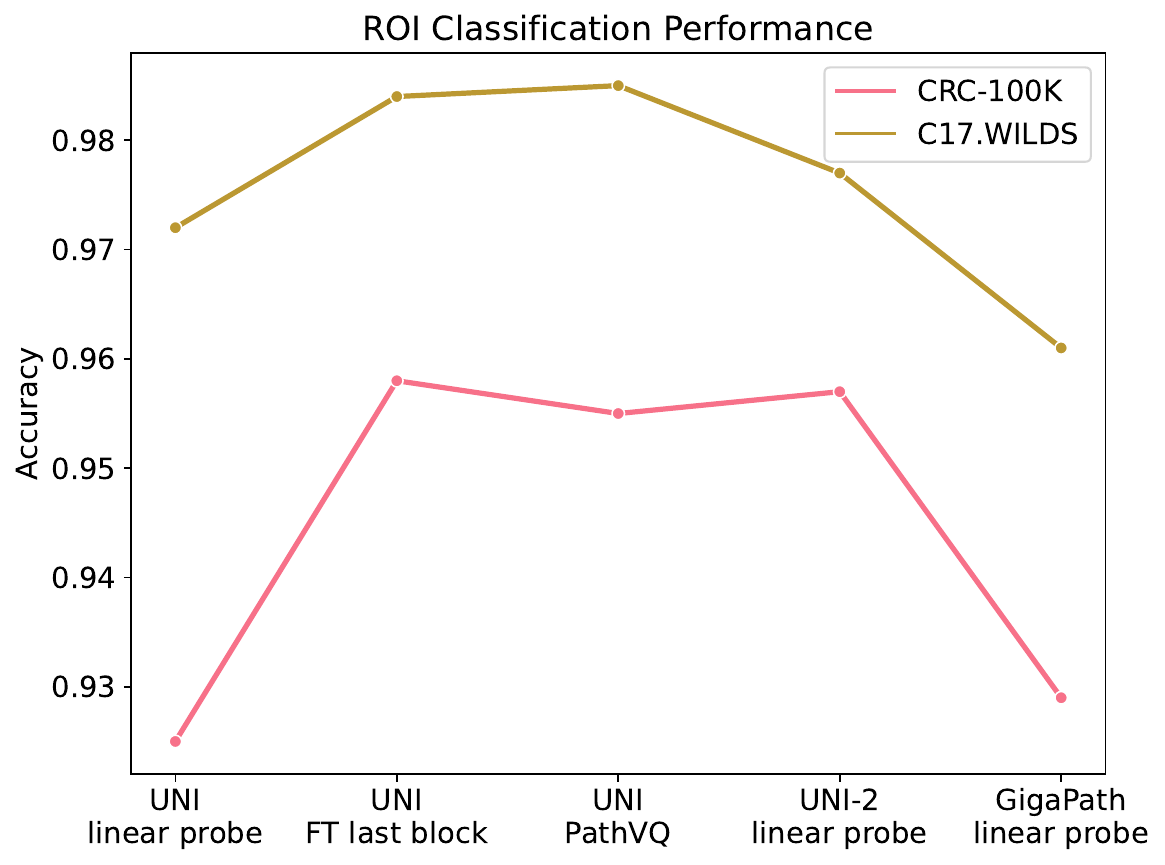} \\
\caption{ROI classification. Obviously, further fine-tuning (FT) on the last block of UNI ViT can further improve the downstream  results. PathVQ, by compressing and reconstructing the patch spatial token, can achieve comparable improvement. UNI-2, however, does not show consistency improvement compared to FT and PathVQ.}
\label{fig:roi}
\end{figure}

\subsubsection{Tile/ROI Classification}

\noindent We evaluate tile/ROI classification performance using dataset CRC-100K~\cite{crc_100k} (9 categories) and Camelyon-17 WILDS~\cite{} tiles (binary). The result in shown in Figure ~\ref{fig:roi}: By only updating the last Transformer block of UNI, the result can be significantly improved. Our PathVQ method is also included and shown comparable improvement to FT. All these results are better than linear probe (freeze backbone and fed [CLS] token feature to classification head). UNI-2, also using linear-probe seems can not scaling up with strong performance on every down-stream tasks.

\subsubsection{WSI Tumor Classification}
\begin{table}[htbp]
  \begin{center}
  \caption{
    Slide-Level Tumor Classification based on CPathFM. The results in the first-row are all trained on UNI, while the second-row we include some recent stronger CPathFMs. The \textcolor{cyan!50}{cyan rows}  are our methods including PathVQ and Slide-level Pre-Training (SPT). The \textcolor{orange!50}{orange rows} demonstrate how much ($\Delta$) of our PathVQ method and UNI-2 improved over UNI with ABMIL setting. The \textbf{bold} and \underline{underline} denote the best and second-best result, respectively. }
  \begin{tabular}{m{3.5cm}<{}||m{1.5cm}<{\centering}m{1.5cm} }
    \midrule[1.2pt]
    & \multicolumn{2}{c}{Tumor classification}\\
    \cline{2-3}
    
      & \multicolumn{2}{c}{\underline{BRACS}}  \\
    Method & F1 & AUC \\
  \midrule
  CLAM-SB~\cite{lu2021data} & 0.640\scriptsize{$\pm$}0.05 & 0.844\scriptsize{$\pm$}0.03 \\
  DTFD-MIL~\cite{Zhang2022DTFDMILDF} &  0.655\scriptsize{$\pm$}0.03 & 0.878\scriptsize{$\pm$}0.02  \\
  TransMIL~\cite{transmil} &  0.592\scriptsize{$\pm$}0.03 & 0.859\scriptsize{$\pm$}0.02 \\
  ABMIL~\cite{abmil} &  0.692\scriptsize{$\pm$}0.03 & 0.875\scriptsize{$\pm$}0.02  \\
  \rowcolor{cyan!20} +PathVQ &  0.730\scriptsize{$\pm$}0.02 & 0.902\scriptsize{$\pm$}0.01  \\
  \rowcolor{orange!20} $\Delta$ over UNI + ABMIL & \footnotesize 0.038 $\uparrow$ & \quad \footnotesize 0.027 $\uparrow$\\
  \rowcolor{cyan!20} +PathVQ + SPT  &  \underline{0.747\scriptsize{$\pm$}0.01} & \underline{0.906\scriptsize{$\pm$}0.01}  \\
  Roformer &  0.678\scriptsize{$\pm$}0.03 & 0.882\scriptsize{$\pm$}0.01  \\
  \rowcolor{cyan!20}+PathVQ & 0.711\scriptsize{$\pm$}0.02 & 0.892\scriptsize{$\pm$}0.01  \\
  \rowcolor{cyan!20}+PathVQ + SPT & \textbf{0.754\scriptsize{$\pm$}0.02}  & \textbf{0.910\scriptsize{$\pm$}0.01} \\

  \midrule
  GigaPath &  {0.677\scriptsize{$\pm$}0.03} & {0.862\scriptsize{$\pm$}0.03} \\
  TITAN & {0.696\scriptsize{$\pm$}0.04} & {0.891\scriptsize{$\pm$}0.01}\\ 
  UNI-2 + ABMIL   & {0.698\scriptsize{$\pm$}0.03} & {0.887\scriptsize{$\pm$}0.02}\\
  \rowcolor{orange!20} $\Delta$ over UNI + ABMIL & \footnotesize 0.006 $\uparrow$ & \quad \footnotesize 0.012 $\uparrow$\\
  
  \midrule[1.2pt]
  \end{tabular}   
   
  \label{tab:wsi_cls} 
  \end{center}   
  \end{table}

We evaluate our method on the BRACS~\cite{brancati2021bracs}, a dataset with three categories—negative, benign, and malignant cancer.
(We notice that popular-used WSI binary tumor classification tasks(e.g. Camelyon ~\cite{camelyon}, TCGA-NSCLC ~\cite{tcga}) are nearly solved (AUC$>$96) given CPathFMs progress. So here we mainly focus on more difficult task, like more categories, and will explore and validate more difficult datasets in near future.)

\noindent \textbf{Compared Baselines:}  
Since our method primarily focuses on extracting improved tile-level features for WSI analysis, we compare it against various WSI analysis models with different architectural designs:
ABMIL~\cite{abmil}, DSMIL~\cite{li2021dual} (introduces a max-pooling branch alongside the attention mechanism), and DTFD-MIL~\cite{Zhang2022DTFDMILDF} (employs sub-bags for hierarchical learning). TransMIL~\cite{transmil} (leveraging Nyström self-attention~\cite{xiong2021nystromformer} for computational efficiency), Transformer with 2-d RoPE~\cite{su2021roformer,li2024rethinking,pochet2023roformer}.

CPathFMs like GigaPath (a 12-layers WSI-Transformer (efficiently implemented using LongViT~\cite{wang2023image}), pretrained on large-scale private data via MAE~\cite{MAE}, with a [CLS] token as tile feature), TITAN~\cite{xu2024whole,ding2024titan} (a 6-layers WSI-Transformer with 2D-ALiBi positional encoding~\cite{press2021alibi, li2024rethinking}, pretrained on large-scale private data using iBOT~\cite{zhou2021ibot}, with Conch-v1.5 as the tile feature extractor ([CLS] token).), and UNI-2 are also included.
Certain works that focus on orthogonal aspects, such as overfitting mitigation, hard instance mining, etc. ~\cite{zhu2024dgr,ACMIL,zhang2024adr,qu2022bidirectional,tang2023multiple,tang2024feature,cui2023bayesmil,lin2023interventional,yang2024mambamil}, are not included in our primary comparison.

For all the experiments, we report the macro-AUC and macro-F1 scores (over five-runs or five-fold cross validation) because of class imbalance.

\noindent \textbf{WSI Classification Results Analysis:}  
The results are reported in Table \ref{tab:wsi_cls}. We can first observe that ABMIL and Roformer show significant improvement when combined with our PathVQ compressor into UNI. The results difference of UNI+PathVQ+ABMIL (about 3\% improvement with adding 1M tile data) and UNI2+ABMIL (about 1\% improvement with adding large-scale ($>>$1M) of tile data, and $2\sim 6\times$ model size) demonstrate that the scalability of previous CPathFMs are bottlenecked by the [CLS] token information losses.
In addition, the results of our slide-level pretraining (\textbf{SPT}) also show consistency improvement compared with random initialization.

\subsubsection{WSI Survival Prediction}

\begin{table*}[!ht]
\centering
\caption{\textbf{Survival prediction} Results of PathVQ and baselines for measuring patient disease-specific survival. All methods in Prototype and MIL use UNI features~\cite{chen2024towards}. Best performance in \textbf{bold}, second best \underline{underlined}. }
\begin{tabular}{ll|c|c|c|c|c}
\toprule

&\textbf{TCGA} & BRCA & CRC & BLCA  & UCEC & KIRC \\
\cline{3-7}

\midrule
\parbox[t]{20mm}{\multirow{3}{*}{\rotatebox[origin=c]{0}{{\makecell{{\textbf{Prototype}}\\\makecell{\scriptsize{} (unsup. UNI)}}}}}}
&\multirow{1}{*}{\color{gray}H2T~\cite{vu2023handcrafted}}  & {\color{gray}0.672}\scriptsize{$\pm$}{\color{gray}0.07}  & {\color{gray}0.639}\scriptsize{$\pm$}{\color{gray}0.11} & {\color{gray}0.566}\scriptsize{$\pm$}{\color{gray}0.05} & {\color{gray}0.715}\scriptsize{$\pm$}{\color{gray}0.09} & {\color{gray}0.703}\scriptsize{$\pm$}{\color{gray}0.11}  \\
&\multirow{1}{*}{\color{gray}OT~\cite{mialon2021a}}  & {\color{gray}0.755}\scriptsize{$\pm$}{\color{gray}0.06}  & {\color{gray}0.622}\scriptsize{$\pm$}{\color{gray}0.09} & {\color{gray}0.603}\scriptsize{$\pm$}{\color{gray}0.04} & {\color{gray}0.747}\scriptsize{$\pm$}{\color{gray}0.08} & {\color{gray}0.695}\scriptsize{$\pm$}{\color{gray}0.09}  \\
&\multirow{1}{*}{\color{gray}PANTHER~\cite{song2024panther}} & {\color{gray}0.758}\scriptsize{$\pm$}{\color{gray}0.06} & {\color{gray}0.665}\scriptsize{$\pm$}{\color{gray}0.10} & {\color{gray}0.612}\scriptsize{$\pm$}{\color{gray}0.07} & {\color{gray}0.757}\scriptsize{$\pm$}{\color{gray}0.10} & {\color{gray}0.716}\scriptsize{$\pm$}{\color{gray}0.10}  \\
\midrule

\parbox[t]{0mm}{\multirow{11}{*}{\rotatebox[origin=c]{0}{{\makecell{{\textbf{MIL}}\\\makecell{\scriptsize{} (supervised. UNI)}}}}}} &
\multirow{1}{*}{AttnMISL~\cite{yao2020whole}} & 0.627\scriptsize{$\pm$}0.08  & 0.639\scriptsize{$\pm$}0.10 & 0.485\scriptsize{$\pm$}0.06 & 0.581\scriptsize{$\pm$}0.12 & 0.649\scriptsize{$\pm$}0.09  \\
&\multirow{1}{*}{ILRA~\cite{xiang2023exploring}} & 0.649\scriptsize{$\pm$}0.10 & 0.555\scriptsize{$\pm$}0.10 & 0.550\scriptsize{$\pm$}0.04 & 0.632\scriptsize{$\pm$}0.02 & 0.637\scriptsize{$\pm$}0.14  \\

&\multirow{1}{*}{TransMIL~\cite{transmil}}  & 0.612\scriptsize{$\pm$}0.07 & \textbf{0.684\scriptsize{$\pm$}0.06} & 0.595\scriptsize{$\pm$}0.06 &  0.695\scriptsize{$\pm$}0.08 & 0.671\scriptsize{$\pm$}0.10  \\
&\multirow{1}{*}{ABMIL~\cite{abmil} $\dagger$}     & 0.644\scriptsize{$\pm$}0.05 & {0.608\scriptsize{$\pm$}0.09} & 0.550\scriptsize{$\pm$}0.06 & 0.669\scriptsize{$\pm$}0.07 & 0.684\scriptsize{$\pm$}0.06  \\

&\multirow{1}{*}{ABMIL reproduce}     & 0.633\scriptsize{$\pm$}0.06 & 0.612\scriptsize{$\pm$}0.08 & 0.540\scriptsize{$\pm$}0.07 & 0.671\scriptsize{$\pm$}0.08 & 0.691\scriptsize{$\pm$}0.08  \\
&\cellcolor{cyan!20}\multirow{1}{*}{+ PathVQ} & \cellcolor{cyan!20}0.655\scriptsize{$\pm$}0.05 & \cellcolor{cyan!20}0.649\scriptsize{$\pm$}0.12 & \cellcolor{cyan!20}\underline{0.608\scriptsize{$\pm$}0.05} & \cellcolor{cyan!20}0.721\scriptsize{$\pm$}0.10 & \cellcolor{cyan!20}0.760\scriptsize{$\pm$}0.08  \\
& \cellcolor{orange!20}\multirow{1}{*}{$\Delta$ over ABMIL} &  \cellcolor{orange!20} \footnotesize 0.022  $\uparrow$ &  \cellcolor{orange!20} \footnotesize 0.037 $\uparrow$ & \cellcolor{orange!20} \footnotesize 0.068 $\uparrow$  & \cellcolor{orange!20} \footnotesize 0.050 $\uparrow$ & \cellcolor{orange!20} \footnotesize 0.069 $\uparrow$ \\
  & \cellcolor{cyan!20}\multirow{1}{*}{+PathVQ + SPT } &  \cellcolor{cyan!20}\textbf{0.674\scriptsize{$\pm$}0.06} &  \cellcolor{cyan!20}0.659\scriptsize{$\pm$}0.08 & \cellcolor{cyan!20}\textbf{0.616\scriptsize{$\pm$}0.05} & \cellcolor{cyan!20}\textbf{0.748\scriptsize{$\pm$}0.11} & \cellcolor{cyan!20}\textbf{0.778\scriptsize{$\pm$}0.08}  \\
 & \multirow{1}{*}{Roformer } & 0.602\scriptsize{$\pm$}0.09  & 0.617\scriptsize{$\pm$}0.13 & 0.572\scriptsize{$\pm$}0.07 & 0.721\scriptsize{$\pm$}0.08 & 0.655\scriptsize{$\pm$}0.13  \\
  & \cellcolor{cyan!20}\multirow{1}{*}{+PathVQ } & \cellcolor{cyan!20}0.644\scriptsize{$\pm$}0.07 & \cellcolor{cyan!20}0.587\scriptsize{$\pm$}0.09 & \cellcolor{cyan!20}0.597\scriptsize{$\pm$}0.05 & \cellcolor{cyan!20}\underline{0.741\scriptsize{$\pm$}0.09} & \cellcolor{cyan!20}0.748\scriptsize{$\pm$}0.09  \\
 & \cellcolor{cyan!20}\multirow{1}{*}{+PathVQ + SPT} & \cellcolor{cyan!20}\underline{0.673\scriptsize{$\pm$}0.07} & \cellcolor{cyan!20}\underline{0.679\scriptsize{$\pm$}0.08} & \cellcolor{cyan!20}0.603\scriptsize{$\pm$}0.05 & \cellcolor{cyan!20}0.734\scriptsize{$\pm$}0.11 & \cellcolor{cyan!20}\underline{0.765\scriptsize{$\pm$}0.08}  \\

\midrule

\parbox[t]{0mm}{\multirow{4}{*}{\rotatebox[origin=c]{0}{{\makecell{{\textbf{CPathFMs}}\\\makecell{\scriptsize{} (SOTA, ckpt-only)}}}}}} &
\multirow{1}{*}{\textcolor{gray}{CHIEF~\cite{wang2024pathology}}} & \textcolor{gray}{0.737\scriptsize{$\pm$}0.04} & \textcolor{gray}{0.680\scriptsize{$\pm$}0.08} & \textcolor{gray}{0.599\scriptsize{$\pm$}0.02} & \textcolor{gray}{0.758\scriptsize{$\pm$}0.10} & \textcolor{gray}{0.736\scriptsize{$\pm$}0.06} \\
& \multirow{1}{*}{\textcolor{gray}{GigaPath~\cite{xu2024whole}}} & \textcolor{gray}{0.687\scriptsize{$\pm$}0.08} & \textcolor{gray}{0.628\scriptsize{$\pm$}0.08} & \textcolor{gray}{0.589\scriptsize{$\pm$}0.05} & \textcolor{gray}{0.779\scriptsize{$\pm$}0.10} & \textcolor{gray}{0.751\scriptsize{$\pm$}0.07} \\
& \multirow{1}{*}{\textcolor{gray}{TITAN~\cite{ding2024multimodal}}} & \textcolor{gray}{0.713\scriptsize{$\pm$}0.04} & \textcolor{gray}{0.710\scriptsize{$\pm$}0.11} & \textcolor{gray}{0.657\scriptsize{$\pm$}0.05} & \textcolor{gray}{0.789\scriptsize{$\pm$}0.09} & \textcolor{gray}{0.774\scriptsize{$\pm$}0.06} \\
& \multirow{1}{*}{\textcolor{gray}{UNI-2 + ABMIL}} & \textcolor{gray}{0.614\scriptsize{$\pm$}0.02} & \textcolor{gray}{0.618\scriptsize{$\pm$}0.11} & \textcolor{gray}{0.539\scriptsize{$\pm$}0.08} & \textcolor{gray}{0.672\scriptsize{$\pm$}0.08} & \textcolor{gray}{0.659\scriptsize{$\pm$}0.11} \\

\bottomrule
\end{tabular}
\label{tab:surv}
\end{table*}

We evaluate survival prediction on five TCGA datasets: BRCA, BLCA, CRC, UCEC, and KIRC. The model is trained using the negative log-likelihood (NLL) loss and evaluated using the c-index with 5-fold cross validation (the result of last epoch is reported).

For fair comparison, we follow the default training pipeline of PANTHER~\cite{song2024panther}, including hyper-parameters and data splits, and integrate our proposed model modifications along with pretrained weights.

\noindent \textbf{Compared Baselines:}  
We categorize the baselines into three groups:
\textbf{Unsupervised Prototype-Based Approaches:} H2T~\cite{vu2023handcrafted} (clusters tile embeddings and pools them within each cluster), OT~\cite{mialon2021a} (aggregates patch features into a set of prototypes using Optimal Transport), PANTHER~\cite{song2024panther} (models prototype tile embeddings via a Gaussian Mixture Model).
\textbf{Supervised MIL Models:}  AttnMISL~\cite{yao2020whole} (combines prototype-based learning with MIL), ABMIL~\cite{abmil}, TransMIL~\cite{transmil}, ILRA~\cite{xiang2023exploring}, and Transformer with RoPE~\cite{su2021roformer}.
\textbf{Slide-Level CPathFMs:} CHIEF~\cite{wang2024pathology}: A large-scale ABMIL-pretrained model using contrastive learning to predict organ source, with CTransPath as the tile feature extractor (mean-pooled features), GigaPath~\cite{xu2024whole} and TITAN~\cite{ding2024titan}. UNI-2 ~\cite{chen2024towards} feature extractor with ABMIL model.

\noindent \textbf{Survival Prediction Results Analysis:}  
The results are reported in Table \ref{tab:surv}. We can observe that ABMIL and Roformer show significant improvement when combined with our PathVQ compressor into UNI. But for Roformer with large-scale of parameters, the performance get easily overfitting to labels thus showing interior performance, which demonstrate the necessity of pretraining. For UNI-2, the improvement is marginal, proving our previous claim on the [CLS] bottleneck of scalability.

\subsection{Ablations and Visualization}
\label{sec:ablation}
\begin{figure}[htbp]
\centering
\includegraphics[width=1.\linewidth]{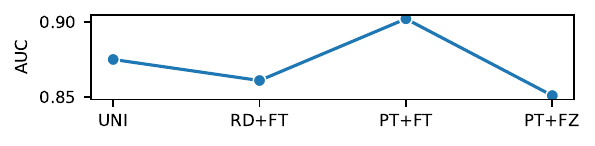} \\
\caption{The PathVQ feature with Convs need pretraining and aligning tile's level-0 feature to attain good feature space and compression. RD: random initialize Convs. FT: fine-tuning during WSI analysis. PT: pretrained during VQ learning (align to level-0 tile feature). FZ: freeze during WSI analysis. }
\label{fig:convs}
\end{figure}

\noindent \textbf{Convs pretraining ablation:} Please check Figure \ref{fig:convs}, experimented on BRACS.

For additional ablation studies and visualizations, please refer to the Appendix. The main findings:

\noindent The VQ reconstruction performance remains relatively stable when varying the quantized embedding dimension (32) and codebook size (16384). 

\noindent The reconstruction results of MSVQ show improvements. 

\noindent The PathVQ can also improve Conch-v1.5.

\section{Conclusion}
In this work, we introduced a novel vector quantization (VQ) distillation framework to address the inherent limitations of existing computational pathology foundation models (CPathFMs) in whole-slide image (WSI) analysis. By efficiently compressing patch-level spatial tokens while preserving critical spatial and contextual information, our method significantly reduces storage and computational costs without compromising performance. Furthermore, our multi-scale VQ (MSVQ) strategy unifies patch- and tile-level features, not only improving feature reconstruction but also serving as an effective self-supervised learning (SSL) supervision target for slide-level pretraining. These advancements enable the extraction of rich and discriminative representations, facilitating more accurate and interpretable predictions for cancer diagnosis and prognosis.

\section{Appendix}
\subsection{Further Related Work on Multiple Instance Learning}
A widely adopted approach is Attention-based Multi-Instance Learning (ABMIL)~\cite{abmil}, which assigns adaptive instance-level weights to focus on the most informative tiles within WSIs. This significantly reduces annotation requirements while enabling meaningful patient-level diagnosis. Building upon this paradigm, advanced weakly-supervised methods such as DS-MIL~\cite{li2021dual}, CLAM~\cite{lu2021data}, and others~\cite{Zhang2022DTFDMILDF,javed2022additive,qu2022bidirectional,cui2023bayesmil,bergner2023iterative} have been proposed. To better capture broader contextual information, multi-scale modeling has been explored to incorporate both fine-grained details and global representations~\cite{chen2022scaling,li2021dual}. Additionally, graph-based models~\cite{chen2021slide,li2018graph,guan2022node,chan2023histopathology,fourkioti2024camil} have been introduced to enhance context-awareness by structuring patch-level relationships within WSIs. Transformer-based architectures~\cite{chen2022scaling,transmil,li2024rethinking} leverage pairwise token interactions to effectively capture contextual dependencies in WSIs.

\subsection{Data Description}

BReAst Carcinoma Subtyping (BRACS)~\cite{brancati2021bracs} collect H$\&$E stained Histology Images, containing 547 WSIs for three lesion types, i.e., benign, malignant and atypical, which are further subtyped into seven categories. Here, since the WSIs number is limited, we only perform three class subtyping.

TCGA ~\cite{tcga}: Breast
Invasive Carcinoma (BRCA, n = 1, 041, WSI =
1, 111), Colon and Rectum Adenocarcinoma (CRC, n =
566, WSI = 575), Bladder Urothelial Carcinoma (BLCA,
n = 373, WSI = 437), Uterine corpus endometrial carcinoma (UCEC, n = 504, WSI = 565), Kidney renal clear
cell carcinoma (KIRC, n = 511, WSI = 517). The
train/val split is performed on the patient level.

\subsection{Experimental settings}
\noindent \textbf{VQ Pretraining:}  
We conduct VQ pretraining on 1M randomly cropped $224 \times 224$ tiles extracted from all TCGA~\cite{tcga} diagnostic pathology WSIs.  
During training, the CPathFM backbone (e.g., UNI with ViT-Large) remains frozen. 
The input tile images are augmented using \texttt{RandomCrop} (minimum ratio: 0.4) and \texttt{RandomHorizontalFlip} (probability: 0.5). 
The codebook has a size of $C=8192$ with an embedding dimension of $16$.
The MLP encoder consists of two linear layers with a \texttt{tanh} activation in between, transforming the feature dimension from $1024$ to $16$.  
The decoder first upsamples the feature dimension from $16$ to $768$ using a linear layer, followed by three Transformer blocks. Another linear layer then maps the features from $768$ to $1024$, ensuring alignment with the original feature tokens.  

For MSVQ, we employ a multi-scale resolution list: $\{1\times1, 2\times2, 4\times4, 7\times7, 14\times14\}$.  

The model is trained on 4 RTX-3090 GPUs for 50 epochs using a batch size of 128 tile images. The total training time is approximately 22 hours.
The learning rate is set to $2\times10^{-4}$ with a 5-epoch warmup, followed by cosine decay to a minimum learning rate of $1\times10^{-5}$. The weight decay is $1\times10^{-4}$, and the AdamW optimizer is used with $\beta$ parameters set to $(0.9, 0.99)$. 

\noindent \textbf{WSI-SSL Pretraining:}  
We crop all TCGA diagnostic WSIs into regions of resolution $3584 \times 3584$, yielding a dataset of approximately 250k regions. To facilitate SSL, a pretrained MSVQ model is used to extract the quantized indices of each tile within a region, requiring only about 65MB for storage.

All pretraining is conducted on 4 GPUs for 20 epochs with an initial learning rate of $5\times10^{-4}$. 
The first 2 epochs serve as a warmup phase, followed by cosine decay to a minimum learning rate of $1\times10^{-5}$. The AdamW optimizer is employed with $\beta$ parameters set to $(0.9, 0.98)$ to ensure fast convergence.

For ABMIL pretraining, a batch size of 64 is used due to the model's simplicity. For WSI-Transformer, the batch size is set to 32, with 96 masked tokens out of 256. The learning objective for both models is formulated as a cross-entropy loss over 8192 categories, with ABMIL additionally utilizing soft targets.

\begin{figure}[htbp]
\centering
\includegraphics[width=1.\linewidth]{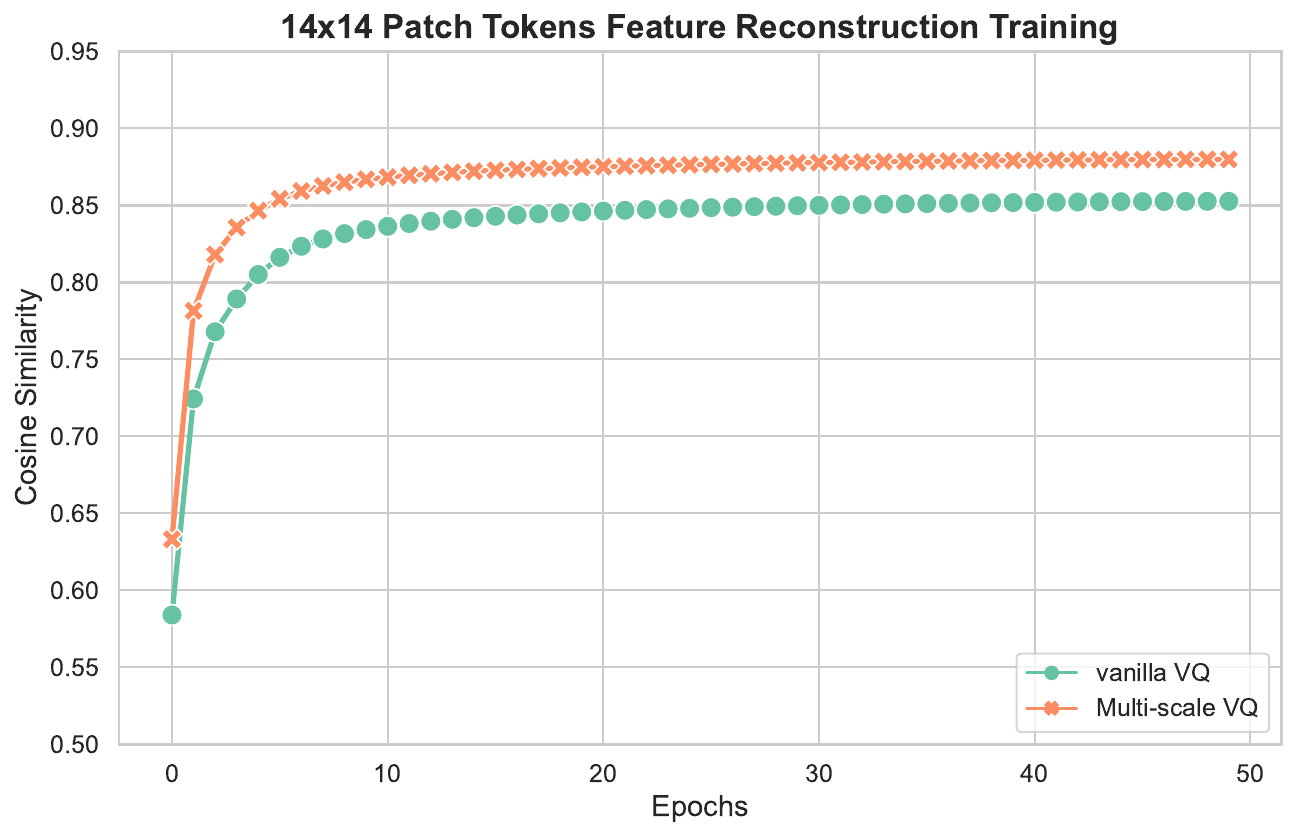} \\
\caption{Reconstruction using Multi-Scale (MSVQ) or not. MSVQ obviously improve the rec performance.}
\label{fig:sup_msvq}
\end{figure}

\begin{figure*}[h]
\centering
\includegraphics[width=.8\linewidth]{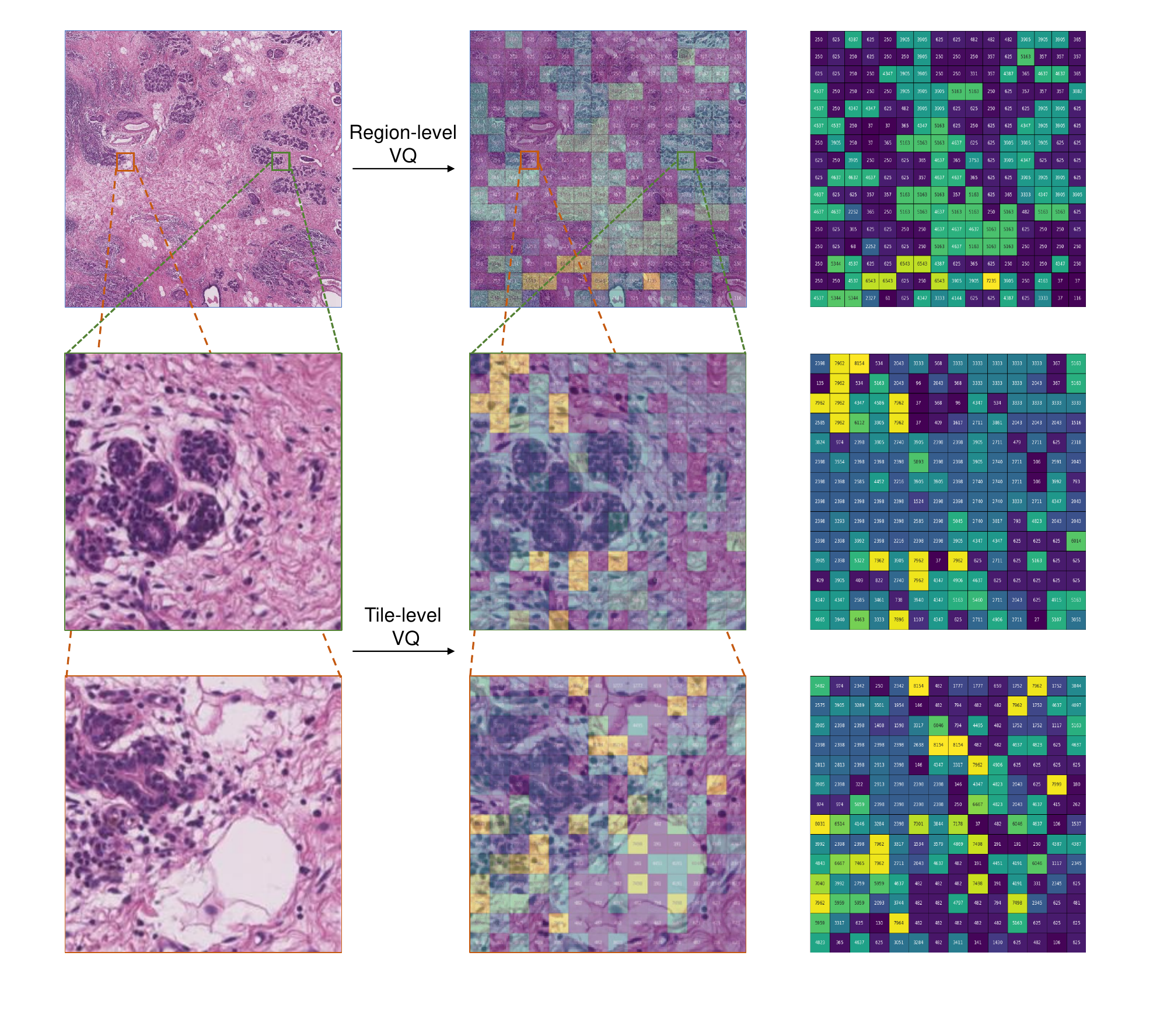} \\
\caption{The VQ are performed on both tile-level and patch-level. The quantized index can be seen as a type of prototypes (n=8k) with strong interpretability. Be aware that since the codebook is too large, the heatmap on different index may share similar color. }
\label{fig_visualize}
\end{figure*}

\begin{figure}[htbp]
\centering
\includegraphics[width=1.\linewidth]{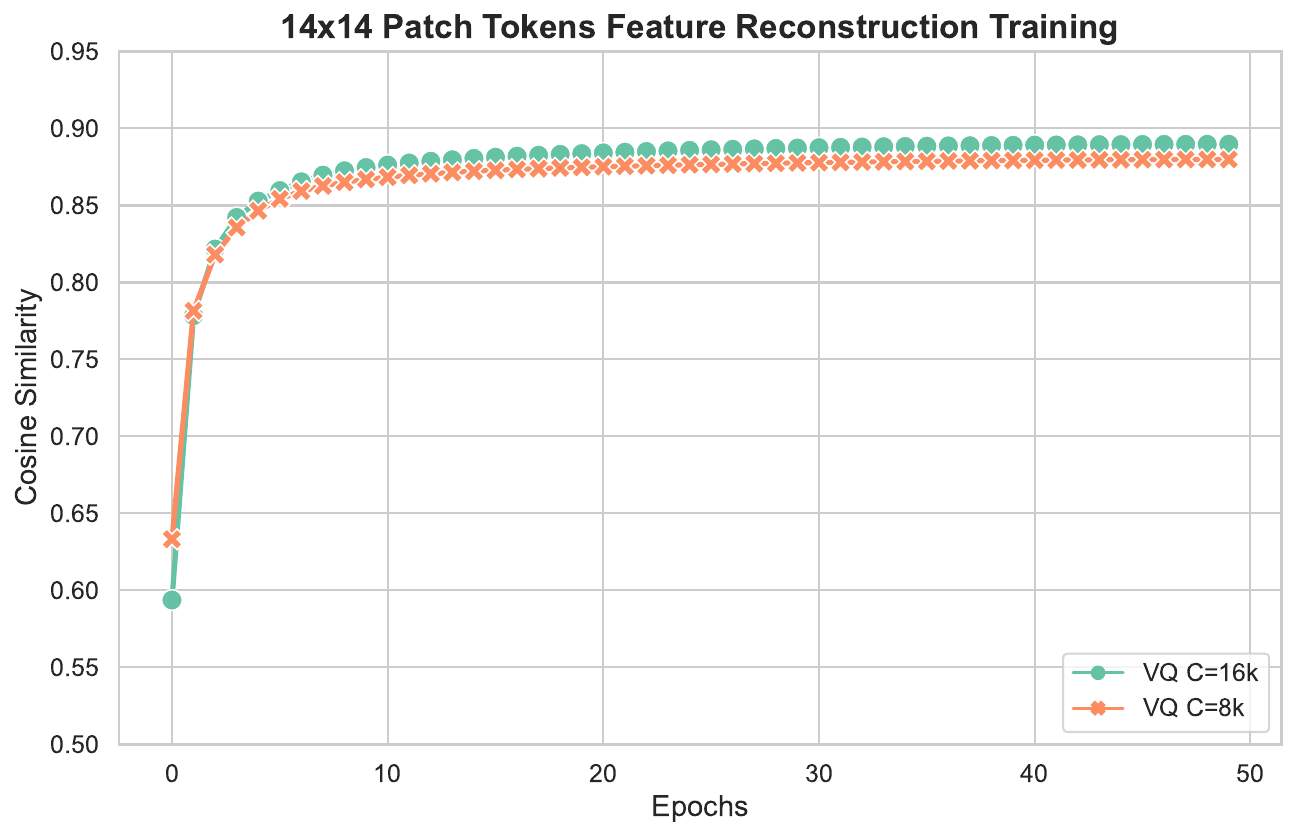} \\
\caption{Reconstruction ablation on quantization codebook size, 8k and 16k.}
\label{fig:supp_codebook}
\end{figure}

\begin{figure}[htbp]
\centering
\includegraphics[width=1.\linewidth]{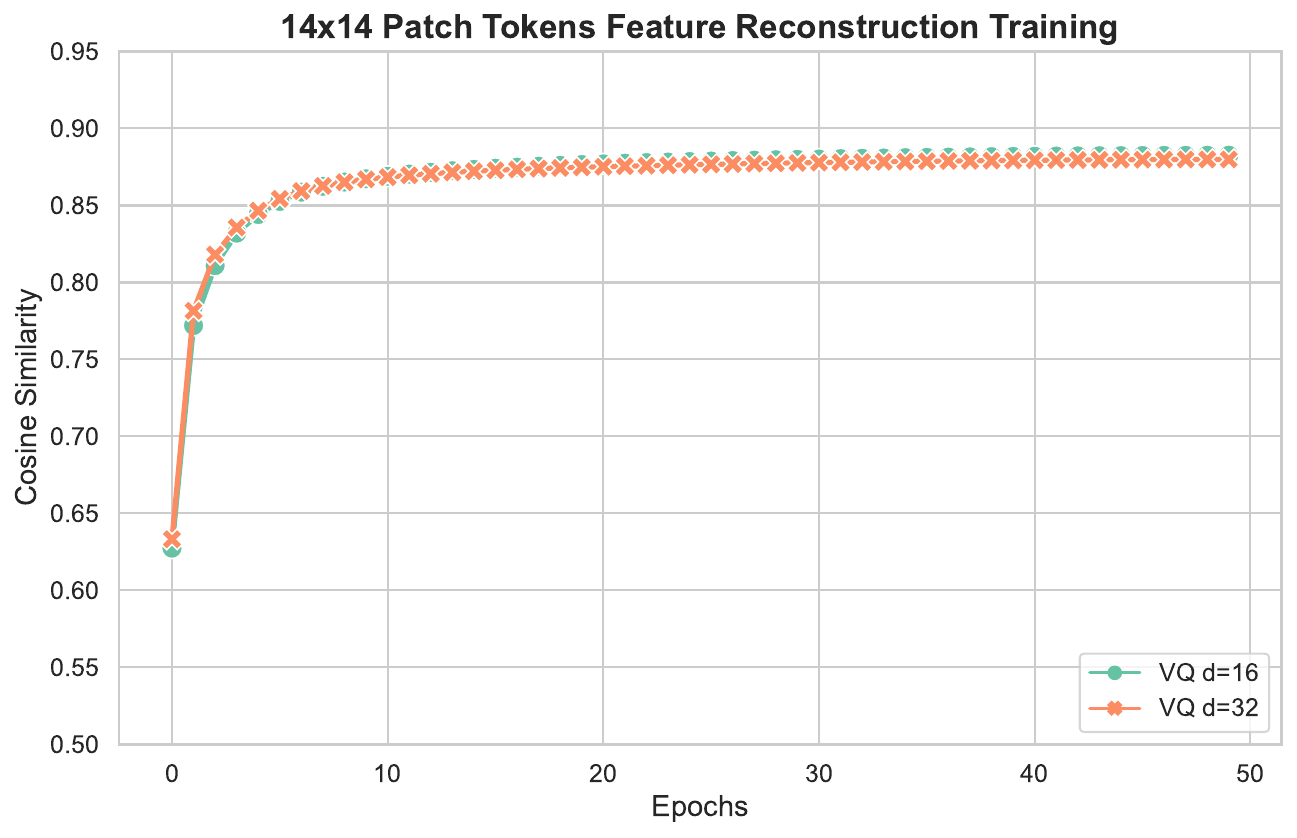} \\
\caption{Reconstruction ablation on quantization codebook embedding, 16 and 32.}
\label{fig:supp_dim}
\end{figure}

\subsection{Ablations on VQ}

\noindent \textbf{Multi-Scale (MSVQ)} Please check Figure \ref{fig:sup_msvq}.

\noindent \textbf{Codebook Size} Please check Figure \ref{fig:supp_codebook}.

\noindent \textbf{Codebook Embedding Dimension} Please check Figure \ref{fig:supp_dim}.

\subsection{Illustrative Visualization}
Please check Figure \ref{fig_visualize}.

\clearpage
{
    \small
    \bibliographystyle{ieeenat_fullname}
    \bibliography{main}
}

\end{document}